\documentclass{article}

\usepackage[preprint]{corl_2026} 
\usepackage{graphicx}
\usepackage{amsmath}
\usepackage{amssymb}
\usepackage{wrapfig}
\usepackage{caption}
\usepackage{authblk}
\usepackage{xspace}
\usepackage{enumitem}
\usepackage{hyperref}
\usepackage{fontawesome5}
\usepackage{bm}
\usepackage{booktabs}
\usepackage{multirow}
\usepackage[table]{xcolor}
\usepackage{pifont}
\usepackage{makecell}
\newcommand{\cmark}{\ding{51}}
\newcommand{\xmark}{\ding{55}}

\title{Towards Precise Intent-Aligned VLA Aerial \\ Navigation via Expert-Guided GRPO}

\author[1,2]{\textbf{Tianyang Chen}$^{*}$}
\author[1,2]{\textbf{Wenjun Li}$^{*}$}
\author[1,2]{\textbf{Xin Zhou}}
\author[1,2]{\textbf{Yuze Wu}$^{\dagger}$}
\author[1,2]{\textbf{Fei Gao}$^{\dagger}$}

\affil[1]{Zhejiang University}
\affil[2]{Differential Robotics}
\affil[ ]{$^{*}$ Equal contribution; {$^{\dagger}$ Corresponding author.}}

\newcommand{\algo}{EG-GRPO\xspace}

\begin{document}
\maketitle


\vspace{-1.0em}

\begin{abstract}
Vision-Language-Action (VLA) models offer a promising end-to-end paradigm for unmanned aerial vehicles (UAVs) to accomplish complex tasks specified by fine-grained instructions. However, standard supervised fine-tuning (SFT) suffers from data scarcity, limited generalization, and weak supervision for nuanced and complicated human intents. Reinforcement fine-tuning offers a natural way to mitigate these challenges and align policy behaviors with human intents through designable feedback, but applying it to aerial navigation remains challenging due to inefficient exploration in expansive continuous spaces. To address these challenges, we introduce an efficient reinforcement learning (RL) framework for VLA-based aerial navigation. At its core, we propose \textbf{\algo} (\textbf{E}xpert-\textbf{G}uided \textbf{G}roup \textbf{R}elative \textbf{P}olicy \textbf{O}ptimization) to augment online rollouts with few-shot expert data. Additionally, we design a heterogeneous pipeline enabling parallel simulation and inference, which reduces rollout time by \textbf{43.5\%}. Across multiple tasks specified by complex human intents, \algo improves the success rate to \textbf{2.13$\times$} that of the SFT baseline, while improving intent alignment performance by \textbf{60.9\%}. These results demonstrate that our framework can move aerial navigation toward precise intent-aligned flight. Our videos are available on \href{https://luoxi-cty.github.io/Intent-Aligned_AerialNav/}{\faGithub\ Intent-Aligned\_AerialNav}, code will be released soon.
\end{abstract}


\section{Introduction}

Vision-Language-Action (VLA) models have emerged as a promising paradigm for autonomous robotic systems~\cite{sapkota2025vision}. By directly mapping high-level natural-language instructions and visual observations to low-level control commands, VLA models provide an end-to-end pipeline that integrates perception, decision-making, and execution. 
However, the prevailing supervised fine-tuning (SFT) framework, which consists of large-scale general pre-training followed by fine-tuning on domain-specific datasets~\cite{tie2025survey}, faces several challenges.
On the one hand, acquiring high-quality datasets, such as complex flight trajectories of unmanned aerial vehicles (UAVs) in navigation scenarios, incurs substantial financial and labor costs~\cite{gao2024efficient}. On the other hand, merely scaling SFT datasets encounters diminishing marginal returns and cannot generalize to out-of-distribution (OOD) scenarios~\cite{liu2025can}.

Recent breakthroughs in VLA models have provided a compelling solution to overcome these bottlenecks. Pioneering studies~\cite{kim2025robot,chen2025tgrpo,ye2025vla} on robotic manipulation have proposed several reinforcement learning (RL) frameworks for VLA models, built upon algorithms such as Proximal Policy Optimization (PPO)~\cite{schulman2017proximal} and Group Relative Policy Optimization (GRPO)~\cite{shao2024deepseekmath}. By leveraging trial-and-error exploration with designable rewards, these works demonstrate that RL can substantially boost performance, adaptability, and generalization of VLA models across diverse scenarios. As manipulation and navigation are core robotic challenges~\cite{wong2025survey}, the success of the VLA-RL framework naturally raises a question: \textbf{\textit{Can RL also strengthen the capability of VLA models in aerial navigation?}}

\begin{figure}[!t] 
    \centering 
    \includegraphics[width=1.0\textwidth]{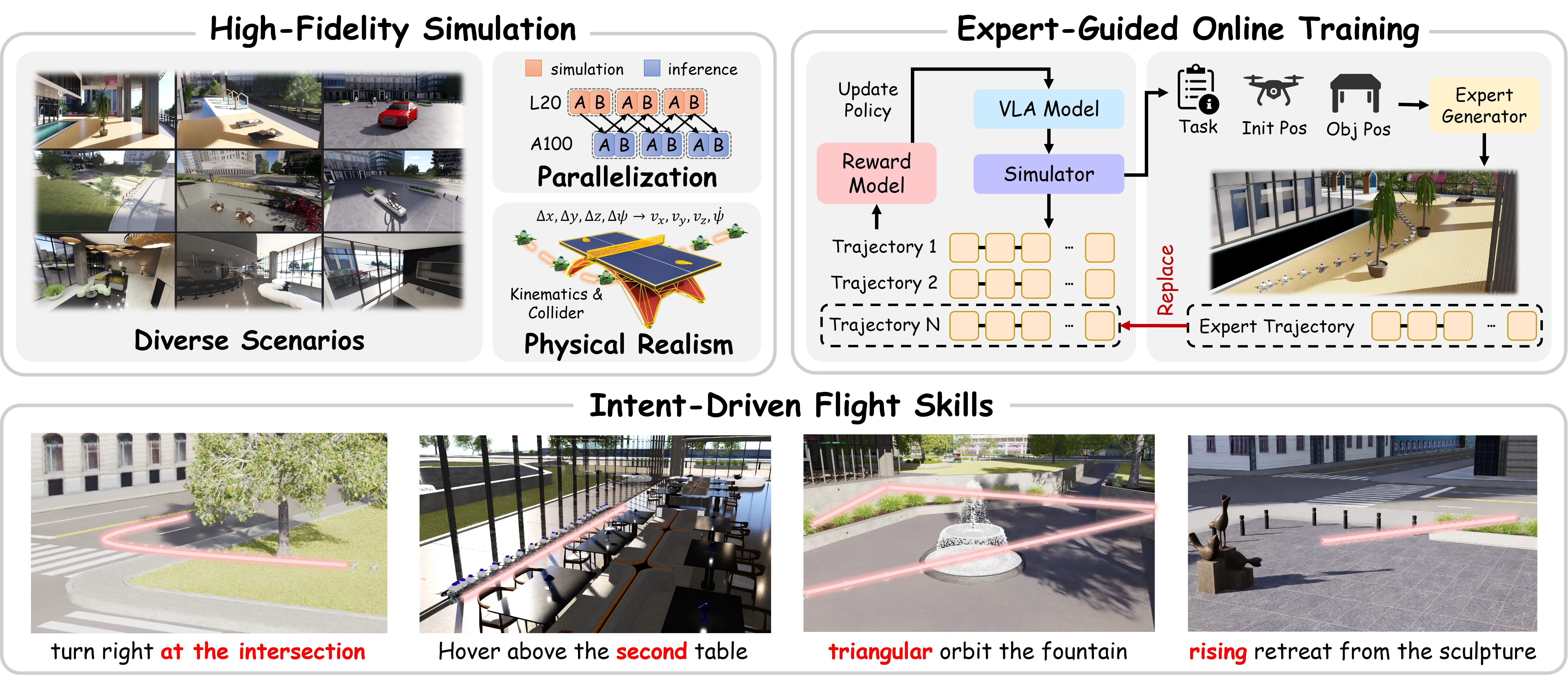} 
    \captionsetup{font={scriptsize}, labelfont={scriptsize}} 
    \caption{\textbf{Overview of our framework.} We introduce a reinforcement learning framework tailored for VLA-based 3D aerial navigation, which integrates a high-fidelity simulation and expert-guided online training to enable UAVs to acquire precise intent-aligned flight skills.} 
    \label{fig:overview} 
\end{figure}

However, directly migrating this framework to aerial navigation is non-trivial due to several domain discrepancies. Manipulation tasks frequently employ outcome-based reward structures (e.g., binary success signals for grasping), focusing primarily on the final state~\cite{luo2025precise,chen2025conrft,kilinc2022reinforcement}. In contrast, navigation is inherently a continuous spatial process that requires evaluating qualitative trajectory attributes at every waypoint~\cite{nahavandi2025comprehensive}---such as safety, smoothness, and alignment with nuanced human intents. Moreover, in 3D UAV navigation, the expansive continuous state space, coupled with the requirements to fulfill fine-grained task instructions, severely constrains the efficiency of random exploration~\cite{liu2023aerialvln,ladosz2022exploration,wu2026precise}. For complex tasks such as “S-shaped bypassing” (bypass the two objects by passing left of the first and right of the second), random exploration struggles to effectively sample high-value behavioral trajectories. 

Beyond domain discrepancies, two challenges hinder practical deployment. First, existing navigation frameworks prioritize long-range trajectory planning~\cite{lin2025openvln,lee2025citynav,jiang2025longfly}, whereas our work focuses on executing complex semantic instructions within confined spaces. This necessitates a simulation infrastructure that integrates high-precision scene modeling, high-fidelity physics, and high-throughput online data acquisition to enable continuous policy learning. Second, the sequential coupling between environment simulation and VLA inference induces substantial resource idling, prolonging data collection cycles and impeding scalable policy iteration~\cite{zang2025rlinf}.

To address these algorithmic and engineering challenges, we introduce a reinforcement learning framework tailored for VLA-based 3D aerial navigation, as illustrated in Fig.~\ref{fig:overview}. To mitigate exploration inefficiency in expansive 3D spaces, we propose \textbf{\algo} (\textbf{E}xpert-\textbf{G}uided \textbf{G}roup \textbf{R}elative \textbf{P}olicy \textbf{O}ptimization), which integrates few-shot expert data into online rollouts. The expert-guided mechanism preserves non-degenerate relative advantages and delivers stable gradient signals that effectively prevent policy stagnation under sparse reward conditions. Concurrently, we deploy a reward model that provides instruction-conditioned feedback for optimization. At the system level, we construct a high-fidelity, interactable UAV simulation platform and develop a heterogeneous pipeline to parallelize simulation and inference, reducing rollout time by \textbf{43.5\%}. Extensive experiments across multiple intent-driven UAV navigation tasks confirm that \algo improves both task completion and intent alignment: compared with the SFT baseline, \algo improves the success rate to \textbf{2.13$\times$} that of the SFT baseline, and improves intent alignment performance by \textbf{60.9\%}. Collectively, these results demonstrate that our framework successfully elevates VLA aerial navigation from basic goal-reaching to precise intent-aligned autonomous flight.

\section{Related Work}

\subsection{Vision-Language Navigation for UAV}

UAV navigation based on human language instructions has evolved from modular language-grounded continuous-control frameworks~\cite{blukis2019learning,blukis2018following,blukis2018mapping} to aerial Vision-Language Navigation (VLN) and VLA models~\cite{wang2024towards,gao2025openfly,wu2025vla}. While these efforts have successfully addressed the challenge of reaching precise endpoints via complex instructions by leveraging large-scale datasets derived from human pilot operations or high-fidelity simulations, they still exhibit certain limitations. Previous research~\cite{zhang2025logisticsvln,saxena2025uav,cai2025flightgpt,sautenkov2025uav} predominantly centers on long-range navigation tasks. These approaches neglect the critical dimension of \textit{\textbf{``how to fly"}}, rendering existing models incapable of executing instructions pertaining to complex and nuanced human intent that extend beyond basic goal-reaching requirements. 

\subsection{Reinforcement Learning for VLA Models}

RL has recently emerged as a promising paradigm for improving VLA models in robotic manipulation~\cite{zhang2025reinbot,chen2025conrft,guo2025improving}. Lu et al.~\cite{lu2025vla} formulate manipulation trajectories as multi-modal multi-turn conversations and use a robotic process reward model to mitigate sparse rewards, enabling online RL fine-tuning of a pretrained autoregressive VLA model. Li et al.~\cite{li2025simplevla} introduce an efficient RL framework with VLA-specific trajectory sampling, scalable parallelization, and multi-environment rendering, showing that binary outcome rewards can improve VLA performance and induce complex behaviors beyond supervised training data. Despite notable progress in developing generalizable and proficient manipulation policies, existing research remains predominantly focused on robotic arm-based manipulation. Although studies such as MoRE~\cite{zhao2025more} and NavGRPO~\cite{li2026trajectory} on quadrupedal robots suggest the potential of VLA-RL in navigation, the specific challenges of aligning aerial trajectories with human instructions remain underexplored. 

\section{Methodology}

\begin{figure}[tbp] 
    \centering 
    \setlength{\belowcaptionskip}{-0pt}
    \includegraphics[width=1.0\textwidth]{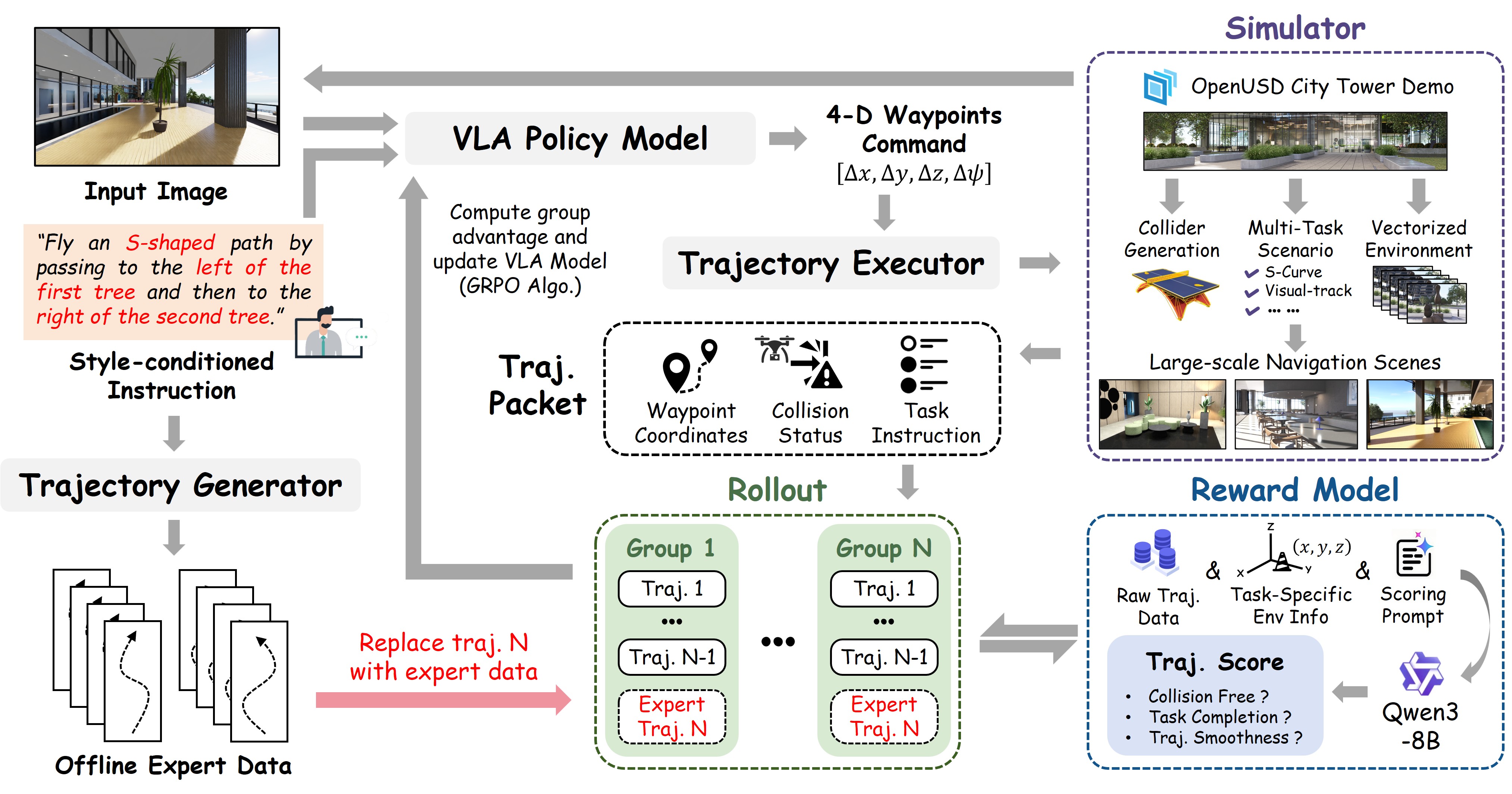} 
    \captionsetup{font={scriptsize}, labelfont={scriptsize}} 
    \caption{\textbf{Overview of the Training Pipeline.} First, the VLA policy generates actions according to visual observations and nuanced instructions. Then, trajectories are executed in parallelized simulations and augmented with offline expert data. Finally, a reward model is deployed to score these trajectories based on task performance, providing fine-grained rewards for updating VLA models.} 
    \label{fig:pipeline} 
\end{figure}

As shown in Fig.~\ref{fig:pipeline}, this section elaborates on the overall architecture and workflow of our framework. 
We build a simulation platform that integrates kinematic simulation with a large collection of high-fidelity scenarios, enabling the policy to generate trajectories in parallelized virtual environments. To mitigate the sparsity of high-value trajectories, we use few-shot expert data to augment online rollouts. Building upon this foundation, we develope a reward model that generates fine-grained trajectory-level reward signals. Finally, the GRPO algorithm completes iterative policy updates with reward signals and action probabilities. Additionally, to address efficiency bottlenecks in large-scale training, we significantly reduced the time overhead of the rollout phase by parallelizing simulation and inference on heterogeneous computing platforms.

\subsection{Problem Formulation and Optimization Objective}


We formulate instruction-conditioned UAV navigation as a trajectory-level policy optimization problem. 
Given a human instruction $\mathcal{I}$ and an observation $o_t$, 
where 
\[
o_t = (o_t^{\mathrm{vis}}, o_0^{\mathrm{vis}}, o_t^{\mathrm{prop}})
\] 
consists of the current RGB observation $o_t^{\mathrm{vis}}$, the initial RGB observation $o_0^{\mathrm{vis}}$, 
and the current proprioceptive state $o_t^{\mathrm{prop}} = [x_t, y_t, z_t, \psi_t]$, 
the VLA policy $\pi_\theta$ autoregressively generates UAV control commands 
$a_t = [\Delta x, \Delta y, \Delta z, \Delta \psi]$. The objective is to maximize the expected trajectory reward $R(\tau, \mathcal{I})$ produced by the reward model, 
where $\tau = \{(o_t, a_t)\}_{t=1}^T$ denotes a complete flight trajectory. 
Formally, the learning goal is to optimize the policy parameters $\theta$ by maximizing the expected return over trajectories conditioned on $\mathcal{I}$:

\begin{equation}
    \theta^* = \arg\max_{\theta} \mathbb{E}_{\tau \sim \pi_\theta(\cdot|\mathcal{I})} \left[ R(\tau, \mathcal{I}) \right],
\end{equation}
By directly optimizing this trajectory-level expectation, the VLA policy learns to generate control sequences that not only achieve the instructed goal but also adhere to nuanced and complex human intents in 3D aerial environments.


\subsection{Policy Initialization with Supervised Fine-Tuning}

For complex flight navigation tasks, the quality of the initial policy directly determines the convergence stability and ultimate performance ceiling during the RL phase. Therefore, acquiring a VLA model with fundamental navigation capabilities is crucial for our pipeline.

We select OpenVLA-OFT~\cite{kim2025fine} as the foundation model due to its strong capabilities and framework compatibility. To adapt it for UAV navigation, we constrain the original action output via a masking function $\mathcal{M}(\cdot)$, filtering out irrelevant degrees of freedom and restricting the output to the 4-DoF pose control commands $[\Delta x, \Delta y, \Delta z, \Delta \psi]$. At the data level, we process the open-source UAV-flow~\cite{wang2026uav} dataset through a custom conversion pipeline, transforming raw flight logs into instruction-conditioned sequential tuples compatible with the training architecture of OpenVLA-OFT. This yields the SFT dataset $\mathcal{D}_{\text{nav}} = \{(o^{(i)}, \mathcal{I}^{(i)}, a^{(i)})\}_{i=1}^N$. The SFT phase then optimizes the policy parameters $\theta$ by minimizing the conditional log-likelihood of trajectories:
\begin{equation}
    \mathcal{L}_{\text{SFT}}(\theta) = -\mathbb{E}_{(o, \mathcal{I}, a) \sim \mathcal{D}_{\text{nav}}} \left[ \log \pi_\theta(a \mid o, \mathcal{I}) \right].
\end{equation}
Minimizing this objective produces the initialized navigation policy $\pi_{init}$. While not globally optimal, $\pi_{init}$ establishes critical behavioral priors that anchor early-stage exploration. 

\subsection{Expert-Guided Group Relative Policy Optimization}

VLA aerial navigation imposes significant data inefficiency due to vast feasible exploration spaces and sparse reward signals. High-value behaviors aligned with nuanced human intents are extremely scarce in purely online rollouts, often causing policy stagnation in low-reward regions. To overcome this, we propose \textbf{\algo}, which seamlessly integrates few-shot expert data into GRPO's group-relative optimization framework. Rather than treating expert data as a separate pre-training phase, we inject it directly into the online learning loop by constructing hybrid trajectory groups that guarantee at least one high-value anchor per update step~\cite{ball2023efficient}.

Formally, during each policy iteration, we sample a trajectory group $\mathcal{G}$ of size $G$ from a hybrid distribution $\mathcal{D}_{\text{mix}} = (1-\rho)\mathcal{D}_{\text{online}} + \rho\mathcal{D}_{\text{expert}}$, where $\rho = 1/G$. In practice, this yields groups composed of $G-1$ online exploration trajectories sampled from $\pi_{\theta_{\text{old}}}$ and exactly one expert trajectory synthesized by a rule-based generator conditioned on task instructions. For such a mixed group $\mathcal{G} = \{\tau_1, \dots, \tau_{\text{expert}}\}$, the group-relative advantage for the $i$-th trajectory is computed as:
\begin{equation}
    \hat{A}_i = \frac{R(\tau_i, \mathcal{I}) - \mu_{\mathcal{G}}}{\sigma_{\mathcal{G}} + \epsilon}, \quad \mu_{\mathcal{G}} = \frac{1}{G}\sum_{j=1}^G R(\tau_j, \mathcal{I}), \quad \sigma_{\mathcal{G}} = \sqrt{\frac{1}{G}\sum_{j=1}^G \left(R(\tau_j, \mathcal{I}) - \mu_{\mathcal{G}}\right)^2},
\end{equation}
where $\epsilon$ ensures numerical stability. Crucially, the expert trajectory $\tau_{\text{expert}} \in \mathcal{G}$ acts as a high-reward anchor that prevents $\mu_{\mathcal{G}}$ from collapsing to zero in highly complex tasks. This guarantees non-degenerate $\sigma_{\mathcal{G}}$ and preserves the discriminative power of $\hat{A}_i$, allowing online trajectories to receive meaningful relative feedback even when they uniformly underperform. The expert trajectory naturally attains a high positive advantage, providing a stable gradient signal for human intent alignment.

The policy parameters $\theta$ are updated by maximizing the clipped surrogate objective over the hybrid group:
\begin{equation}
    \mathcal{L}_{\text{EG-GRPO}}(\theta) = \mathbb{E}_{\mathcal{G} \sim \mathcal{D}_{\text{mix}}} \left[ \frac{1}{G} \sum_{i=1}^G \frac{1}{|\tau_i|} \sum_{t=1}^{|\tau_i|} \min\left( r_{i,t}(\theta) \hat{A}_i, \text{clip}(r_{i,t}(\theta), 1-\delta, 1+\delta) \hat{A}_i \right) \right],
\end{equation}
where $r_{i,t}(\theta) = \pi_\theta(a_{i,t} | o_{i,t}, \mathcal{I}) / \pi_{\theta_{\text{old}}}(a_{i,t} | o_{i,t}, \mathcal{I})$ denotes the step-wise policy ratio, and $\delta$ controls the clipping range. Furthermore, we remove the KL divergence regularization to avoid constraining the exploration of novel behaviors~\cite{li2025simplevla}. By optimizing over $\mathcal{G}$, the algorithm simultaneously learns from online exploration and expert data. This unified formulation resolves the reward sparsity bottleneck: the expert anchor prevents exploration collapse and consistently elevates navigation quality on complex fine-grained instructions, as empirically validated in our experiments.

\subsection{UAV Simulation Platform for VLA-RL}

To support high-throughput policy rollouts and ensure physical realism during online RL training, we construct a vectorized simulation platform based on Isaac Lab~\cite{mittal2025isaac}. As illustrated in Fig.~\ref{fig:pipeline}, raw 3D assets are first processed to generate precise collision meshes, guaranteeing accurate physics-based interaction and safety-critical collision detection. Instruction-conditioned task templates (e.g., target orbiting, obstacle passing, and approach maneuvers) are then configured and instantiated into $N$ parallel environment instances. To bridge policy outputs with physical execution, we integrate a UAV kinematic model that maps the VLA's action deltas directly to continuous, dynamically feasible flight trajectories. This vectorized architecture enables simultaneous trajectory sampling across large-scale scenarios, sustaining the continuous, high-frequency data acquisition required for stable policy iteration.

\subsection{Efficient Rollout with Heterogeneous Parallelization}

In large-scale online RL training, the rollout phase often constitutes the primary time bottleneck. The Isaac Lab simulation platform relies on GPUs equipped with RT Cores (e.g., NVIDIA L20) for hardware-accelerated ray tracing and physical computations, whereas VLA model inference achieves optimal efficiency on compute-optimized accelerators (e.g., NVIDIA A100). Traditional synchronous workflows enforce a strict serial dependency: generating observations, inferring actions, and executing physics steps must occur sequentially. This conventional pipeline forces the RT Core-equipped GPUs and the compute accelerators to strictly alternate between active and idle states, leading to severe hardware underutilization and extended training times~\cite{zang2025rlinf}.

\begin{figure}[tbp] 
    \centering 
    \includegraphics[width=1.0\textwidth]{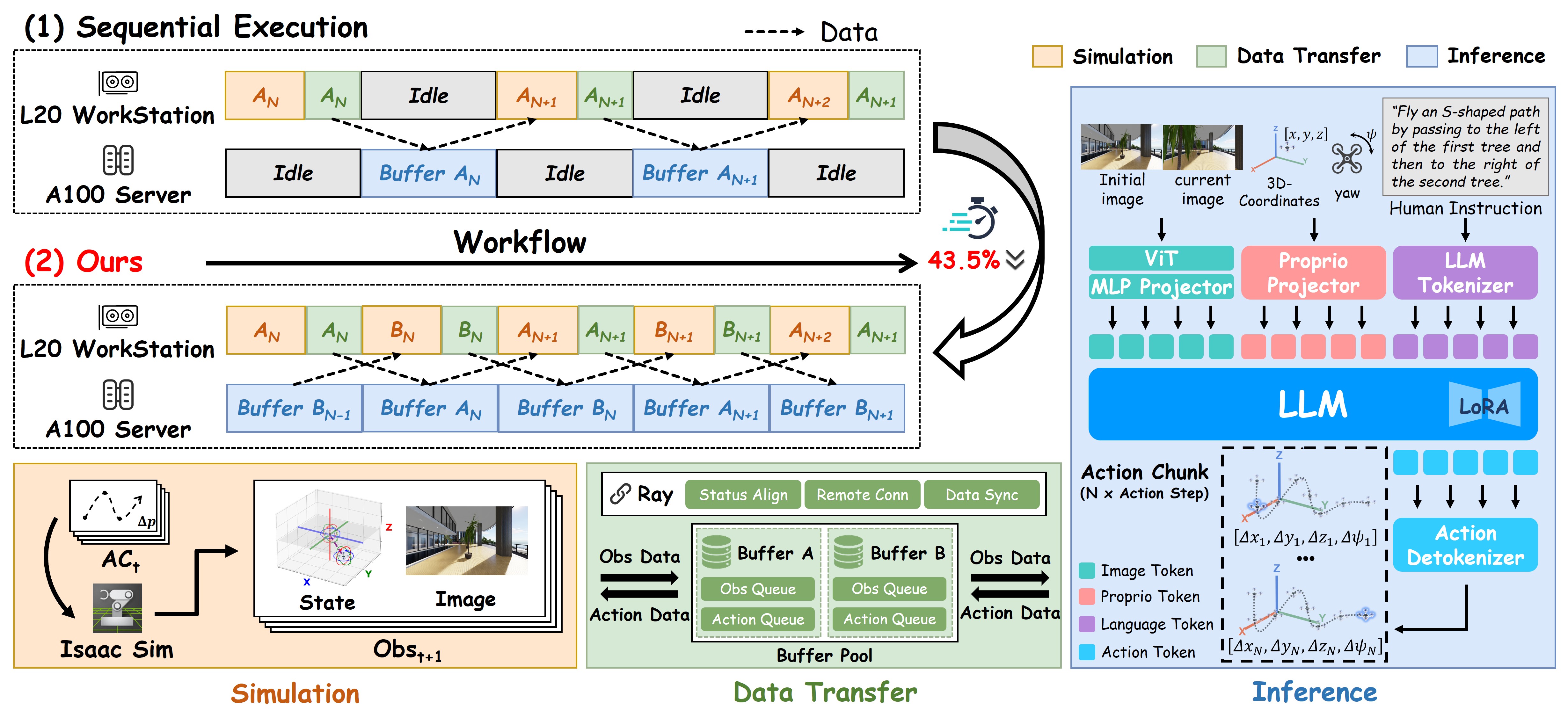} 
    \captionsetup{font={scriptsize}, labelfont={scriptsize}} 
    \caption{\textbf{Heterogeneous Parallelization Workflow.} Conventional serial execution alternates between simulation on RT-core GPUs (L20) and VLA inference on compute GPUs (A100), causing severe hardware idling. Our framework decouples these stages via a dual-group scheduling strategy with a double-buffer mechanism, reducing rollout time by \textbf{43.5\%}.} 
    \label{fig:sim-gen} 
\end{figure}

To circumvent this bottleneck, we design a heterogeneous simulation-inference architecture built upon a double-buffer paradigm. Cross-node data synchronization is facilitated by SSH tunneling and the Ray distributed framework. As depicted in Fig.~\ref{fig:sim-gen}, the environment interaction loop is decoupled into two asynchronous task streams (Group A and Group B) that execute in a cyclic pipeline:

\begin{enumerate}[label={}, leftmargin=2em]
    \item \textbf{Step I:}
    Group A performs physics simulation on the L20 workstation to collect observations, while Group B concurrently runs VLA inference on the A100 server to predict actions.
    
    \item \textbf{Step II:}
    Group A streams its newly collected observations to the A100 server. Simultaneously, Group B transmits its predicted actions back to the L20 workstation.
    
    \item \textbf{Step III: }
    Group A initiates inference on the A100 server using the newly received observations, while Group B executes the received actions within the L20 simulation to collect the next set of observations.
\end{enumerate}
Iterating through \textbf{Steps I–III} sustains a continuous, non-blocking data collection pipeline. This orchestrated scheduling strategy fully exploits heterogeneous compute capabilities, dramatically compressing GPU idle windows. Empirical evaluations confirm that the proposed architecture substantially reduces rollout phase duration, thereby establishing a practically viable engineering pathway for large-scale online RL training.

\subsection{Instruction-Conditioned Reward Model}

To supply trajectory-level optimization signals beyond binary success or failure, we employ an instruction-conditioned reward model that evaluates the quality of complete flight trajectories. Given a flight trajectory $\tau$ and the corresponding instruction $\mathcal{I}$, the reward model assigns a scalar score by assessing  the degree to which the trajectory aligns with the fine-grained instruction. This enables richer feedback for instruction-conditioned UAV navigation, where intermediate trajectory quality, safety, and motion style are often as important as reaching the final goal. Due to the diversity of navigation tasks and the complexity of nuanced instructions, we implement the reward model with a large language model (LLM), which holistically evaluates the trajectory with respect to the given instruction:
\begin{equation}
    R(\tau, \mathcal{I}) = f_{\text{LLM}}(\tau, \mathcal{I}),
\end{equation}
where $f_{\text{LLM}}$ denotes the instruction-conditioned scoring function. To examine the reliability of the evaluator, we validate its scores on 10K trajectories from the online rollouts using a structured human-in-the-loop protocol. Certified drone pilots independently assess the quality of each trajectory with respect to the corresponding instruction, and their judgments are compared against the evaluator outputs. The evaluator achieves a high human agreement rate, suggesting strong consistency with expert assessments. Here, human agreement rate denotes the fraction of samples for which the evaluator's score falls within a predefined tolerance of the averaged pilot assessment. This reward model functions as a plug-and-play scoring component within our framework, providing continuous, instruction-aware feedback to guide policy updates.

\begin{figure}[h]
    \centering
    \includegraphics[width=0.96\textwidth]{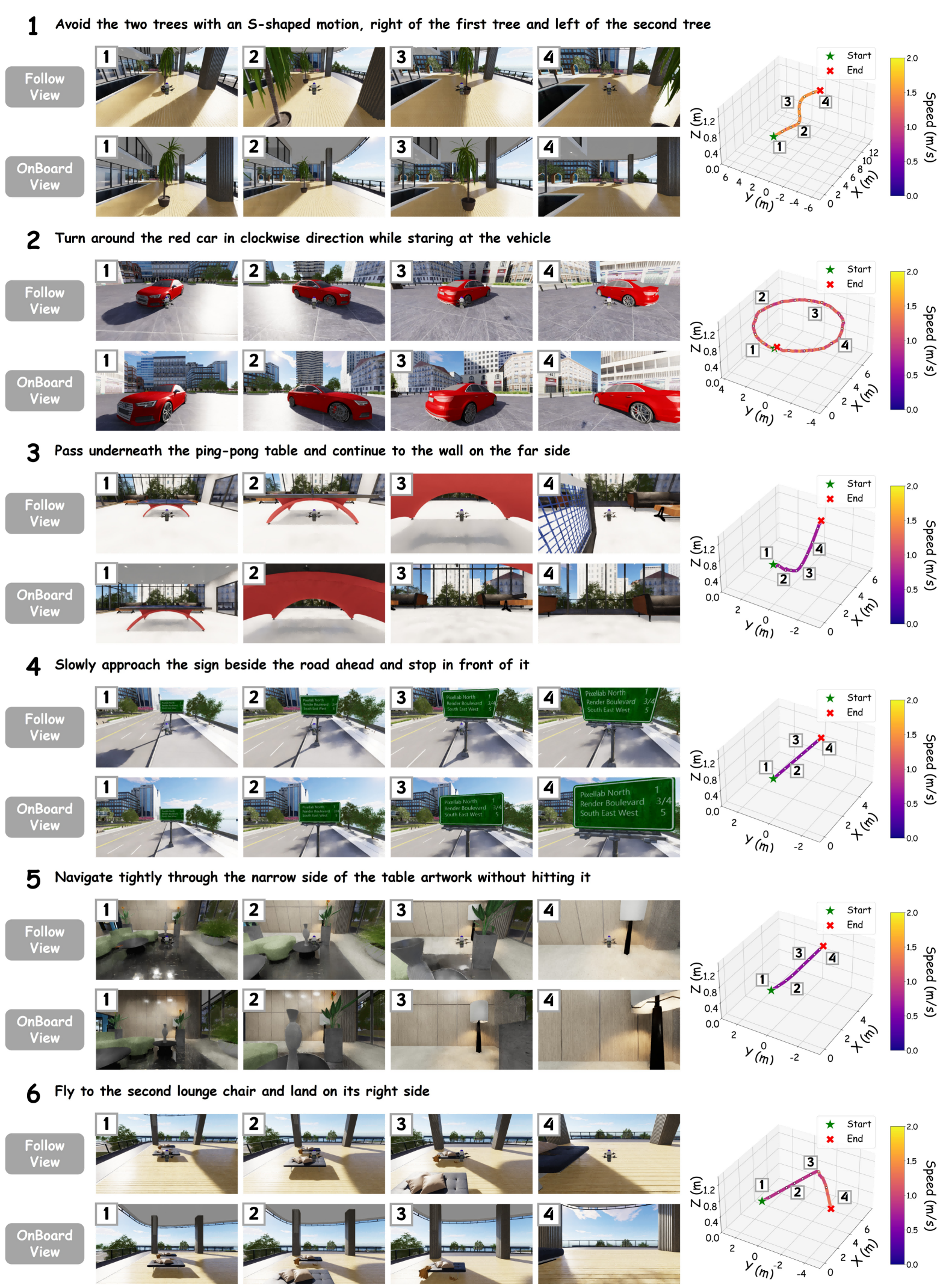}
    \captionsetup{font={scriptsize}, labelfont={scriptsize}} 
    \caption{\textbf{Simulation results of several intent-aligned tasks.} We visualize representative tasks covering S-shaped bypassing, object orbiting, under/over passing, speed-modulated approach, risk-aware bypassing, and precision landing. Each example includes follow-view frames, onboard-view frames, and the corresponding 3D trajectory colored by flight speed.}
    \label{fig:sim_result}
\end{figure}

\section{Experiments}

\subsection{Experimental Setup and Evaluation Metrics}

To evaluate our VLA policy, we construct the \textit{Human-Intent-Driven Navigation} evaluation suite in our simulation platform. Diverse navigation tasks are organized into two task groups according to whether the required behavior is supported by the initial navigation prior. \textbf{Easy tasks} are derived from common navigation instruction templates with additional lightweight intent constraints (e.g., approaching a target quickly, and passing an obstacle from safer side). These tasks evaluate whether our method can improve the navigation performance when the initial policy already possesses reasonable navigation priors. \textbf{Difficult tasks} contain complex fine-grained flight skills that are rarely covered by the original instruction set (e.g., S-shaped bypassing, and under/over passing). These tasks are used to evaluate whether EG-GRPO can elicit complex skills when effective trajectories are scarce with initial policy.

We evaluate task performance using \textbf{Success Rate (SR)} and \textbf{Intent Alignment Score (IAS)}. SR measures whether the UAV successfully completes the given navigation task, while IAS evaluates the degree to which the executed trajectory satisfies the fine-grained human instruction. Specifically, IAS is derived from the trajectory-level score produced by our instruction-conditioned evaluator, which is validated against expert UAV pilot assessments to ensure consistency with human judgment. Compared with SR, which provides a binary measure of task completion, IAS offers a more fine-grained assessment of instruction-following quality.

\subsection{Simulation Results}

We benchmark our policy against two representative vision-language-action models, $\bm{\pi_0}$~\cite{black2024pi_0} and \textbf{OpenVLA-OFT}~\cite{kim2025fine}. Since raw VLA checkpoints do not possess inherent aerial navigation capabilities, we fine-tune both baselines on the UAV-flow dataset using supervised learning. In particular, the SFT-initialized OpenVLA-OFT is used as our policy baseline before RL optimization. 

\begin{table}[h]
\centering
\caption{Results of benchmark evaluations across different task difficulty.}
\label{tab:task_group_results}
\resizebox{\textwidth}{!}{
\begin{tabular}{l|cc|cc|cc}
\toprule
\multicolumn{1}{c|}{\multirow{2}{*}{\textbf{Model}}}
& \multicolumn{2}{c|}{\textbf{Easy}} 
& \multicolumn{2}{c|}{\textbf{Difficult}} 
& \multicolumn{2}{c}{\textbf{Overall}} \\
\cmidrule(lr){2-3} \cmidrule(lr){4-5} \cmidrule(lr){6-7}
& \textbf{IAS Mean $\Uparrow$} & \textbf{SR/\% $\Uparrow$} 
& \textbf{IAS Mean $\Uparrow$} & \textbf{SR/\% $\Uparrow$} 
& \textbf{IAS Mean $\Uparrow$} & \textbf{SR/\% $\Uparrow$} \\
\midrule
$\pi_0$ 
& 6.15 & 41.2 
& 4.47 & 22.8
& 5.31 & 32.0 \\
\cmidrule(lr){1-7}
OpenVLA-OFT 
& 5.19 & 33.5
& 3.81 & 18.7 
& 4.50 & 26.1 \\
\quad w/ \textbf{Ours}
& \textbf{8.28} & \textbf{68.2} 
& \textbf{6.20} & \textbf{43.1} 
& \textbf{7.24} & \textbf{55.6} \\
\rowcolor{pink!18}
\quad $\Delta$
& \textcolor{red}{+3.09} & \textcolor{red}{+34.7}
& \textcolor{red}{+2.39} & \textcolor{red}{+24.4}
& \textcolor{red}{+2.74} & \textcolor{red}{+29.5} \\
\bottomrule
\end{tabular}
}
\end{table}

As shown in Tab.~\ref{tab:task_group_results}, our method substantially improves over both VLA baselines across easy and difficult task groups. On easy tasks, our policy achieves a \textbf{34.7\%} gain in SR over OpenVLA-OFT, improving from \textbf{33.5\%} to \textbf{68.2\%}. Meanwhile, IAS increases from \textbf{5.19} to \textbf{8.28}, indicating stronger intent alignment under common navigation instructions.

The advantage remains clear on difficult tasks. Our method achieves a \textbf{24.4\%} gain in SR over OpenVLA-OFT, improving from \textbf{18.7\%} to \textbf{43.1\%}, while IAS increases from \textbf{3.81} to \textbf{6.20}. These results suggest that the proposed framework is not limited to refining existing navigation behaviors, but also helps the policy acquire complex flight skills under sparse initially effective trajectories. Overall, \algo achieves a \textbf{29.5\%} gain in SR over OpenVLA-OFT, improving from \textbf{26.1\%} to \textbf{55.6\%}, while IAS increases from \textbf{4.50} to \textbf{7.24}. These findings demonstrate that \algo improves instruction-conditioned trajectory quality in diverse UAV navigation tasks.

\subsection{Real-World Deployment}

\begin{figure}[t]
    \includegraphics[width=1.0\textwidth]{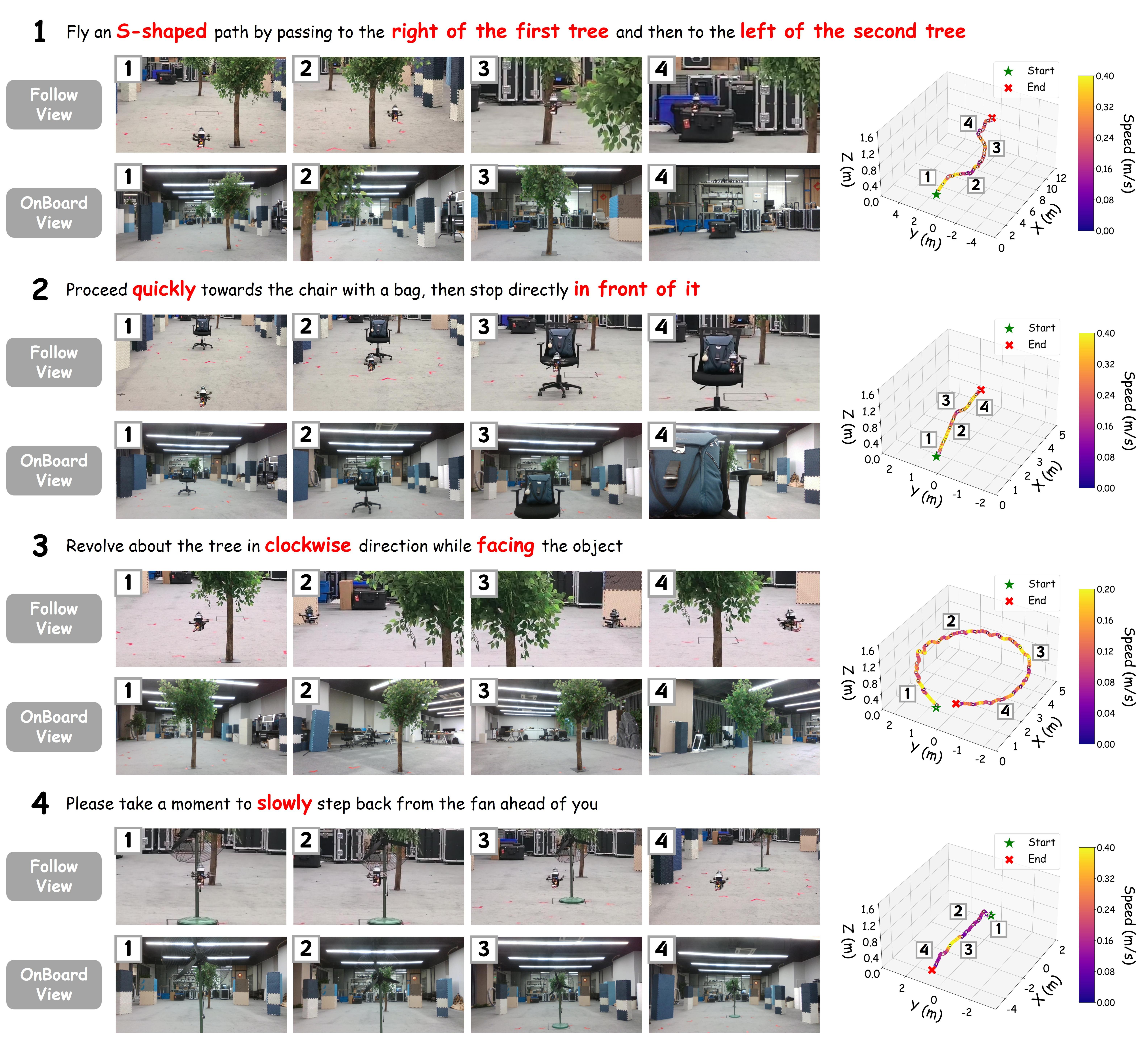}
    \captionsetup{font={scriptsize}, labelfont={scriptsize}} 
    \caption{\textbf{Real-world deployment results.} We show several distinct linguistic commands executed in a physical environment. The results demonstrate zero-shot transfer from simulation to real-world UAV navigation under complex human instructions.}
    \label{fig:realworld_result}
\end{figure}

As shown in Fig.~\ref{fig:realworld_result}, we deploy the trained policy in several representative real-world scenarios to evaluate zero-shot transfer. The tasks involve diverse human-intent constraints, including S-shaped bypassing, speed-modulated object navigation, and visual-locking orbiting. Without additional real-world fine-tuning, the policy successfully executes the given instructions and produces physically feasible trajectories, demonstrating effective alignment between high-level semantic intent and low-level UAV kinematic control.

\subsection{Ablation Studies}

We conduct ablation studies to isolate the contributions of the key components in our framework. In particular, we focus on whether EG-GRPO is necessary for difficult tasks, where high-quality online trajectories are sparse and pure RL exploration may provide unstable optimization signals. For policy learning, we compare three training configurations:

\begin{enumerate}[label={}, leftmargin=2em]
    \item \textbf{Baseline}: the supervised fine-tuned OpenVLA-OFT policy trained on the UAV-flow dataset.
    \item \textbf{Baseline+GRPO}: RL initialized from the SFT baseline and optimized with intent-aware trajectory-level feedback, without expert data injection.
    \item \textbf{Baseline+EG-GRPO}: our complete framework, which injects few-shot expert data into online rollouts during GRPO optimization.
\end{enumerate}

As shown in Tab.~\ref{tab:ablation_results}, only applying GRPO improves the IAS from \textbf{3.81} to \textbf{4.66} and SR from \textbf{18.7\%} to \textbf{26.4\%} in difficult tasks. This confirms that GRPO can refine the SFT policy beyond supervised imitation, even when the initial policy provides limited effective priors for difficult tasks. However, the improvement remains limited, indicating that standard online RL still struggles to discover high-quality trajectories in difficult UAV navigation tasks.

\begin{table}[h]
\centering
\caption{Ablation results of different training components on difficult tasks.}
\label{tab:ablation_results}
\resizebox{\textwidth}{!}{
\begin{tabular}{lccccc}
\toprule
\makecell{\textbf{Method}}
& \makecell{\textbf{SFT with}\\\textbf{Offline Dataset}}
& \makecell{\textbf{RL with}\\\textbf{Trajectory-level Reward}} 
& \makecell{\textbf{Expert Data Injection}\\\textbf{into Online Rollouts}} 
& \makecell{\textbf{IAS Mean $\Uparrow$}} 
& \makecell{\textbf{SR/\% $\Uparrow$}} \\
\midrule
\makecell{Baseline}
& \cmark & \xmark & \xmark
& 3.81 & 18.7 \\
\midrule
\makecell{Baseline\\+GRPO} 
& \cmark & \cmark & \xmark 
& 4.66 & 26.4 \\
\midrule
\makecell{Baseline\\+EG-GRPO} 
& \cmark & \cmark & \cmark 
& 6.20 & 43.1 \\
\bottomrule
\end{tabular}
}
\end{table}

By contrast, EG-GRPO further improves the difficult-task IAS to \textbf{6.20} and SR to \textbf{43.1\%}. Compared with GRPO, expert-guided optimization brings an additional gain of \textbf{1.54} in IAS and \textbf{16.7\%} in SR. This validates the necessity of expert data injection in challenging tasks. By introducing few-shot expert trajectories into online rollout groups, EG-GRPO preserves non-degenerate relative advantages, and stabilizes policy updates when purely online rollouts rarely sample high-quality behaviors.

\begin{wraptable}{r}{0.3\textwidth}
\centering
\vspace{-8pt}
\captionsetup{font={scriptsize}, labelfont={scriptsize}}
\caption{Comparison of the total rollout time cost per RL training step between the sequential baseline and our method.}
\label{tab:rollout_speedup}
\resizebox{\linewidth}{!}{
\begin{tabular}{lcc}
\toprule
\multirow{2}{*}{\textbf{Method}} & \multicolumn{2}{c}{\textbf{Time Cost (s)}} \\
\cmidrule(lr){2-3}
& \textbf{Avg.} & \textbf{Std.} \\
\midrule
Baseline & 904.67 & 58.55 \\
Ours & 511.01 & 91.90 \\
\bottomrule
\end{tabular}
}
\vspace{-10pt}
\end{wraptable}

We further evaluate the efficiency gain brought by the proposed heterogeneous rollout parallelization. As shown in Tab.~\ref{tab:rollout_speedup}, the parallelized simulation-inference pipeline reduces the rollout time from \textbf{904.67s} to \textbf{511.01s} per RL training step. This corresponds to a reduction of approximately \textbf{43.5\%}, or equivalently reduces the rollout time to \textbf{56.5\%} of the sequential baseline. The result verifies that decoupling environment simulation and VLA inference effectively alleviates hardware idling between RT-core GPUs used for physics simulation and compute GPUs used for policy inference. By enabling the two stages to proceed asynchronously through double-buffer scheduling, the proposed pipeline substantially improves rollout throughput and makes online RL fine-tuning more practical for scalable training.

\section{Conclusion}

In this work, we introduce an RL framework for VLA-based UAV navigation under complex, fine-grained language instructions. To overcome inefficient exploration in continuous 3D spaces, we propose \textbf{\algo}, which augments online rollouts with few-shot expert trajectories and uses trajectory-level reward feedback to guide policy updates. We also design a heterogeneous simulation-inference pipeline that reduces rollout time by \textbf{43.5\%}. Experiments across multiple UAV navigation tasks show that \algo improves the success rate from \textbf{26.1\%} to \textbf{55.6\%} over the SFT baseline, while improving intent alignment performance by \textbf{60.9\%}. These results suggest that our framework enables VLA aerial navigation to move beyond goal-reaching toward reliable instruction-aligned autonomous flight.


\clearpage


\bibliography{corl_2026_VLA_RL/reference}  

\newpage

\appendix

\end{document}